\RequirePackage{amsthm}  
\documentclass[pdflatex,sn-mathphys-num]{sn-jnl}
\usepackage{graphicx}
\graphicspath{{../02_Figures}}
\usepackage{todonotes}
\usepackage{soul}
\usepackage{amsmath}
\usepackage{color}
\usepackage{booktabs}
\usepackage{multirow}
\usepackage{graphicx}%
\usepackage{amsthm}%
\usepackage{amssymb}
\usepackage{mathrsfs}%
\usepackage[title]{appendix}%
\usepackage{xcolor}%
\usepackage{textcomp}%
\usepackage{manyfoot}%
\usepackage{booktabs}%
\usepackage{algorithm}%
\usepackage{algorithmicx}%
\usepackage{listings}%
\usepackage{glossaries}
\usepackage{breakurl}
\usepackage{url}
\usepackage{lmodern}
\usepackage{anyfontsize}
\usepackage{hyperref}
\usepackage{upgreek,textgreek}
\usepackage{alphabeta}
\usepackage{csquotes}
\usepackage{orcidlink}
\newcommand*{\figreft}[1]{Figure~\ref{#1}}

\newcommand{\latZ}{\mathcal{Z}}
\newcommand{\latW}{\mathcal{W}}
\newcommand{\scriptN}{\mathcal{N}}
\newcommand{\del}{\partial}

\newcommand{\partialderivative}[2]{\ensuremath{\frac{\del #1}{\del #2}}}
\newcommand{\unit}[1]{\,\text{#1}}

\newcommand{\li}{\left(}
\newcommand{\re}{\right)}
\newacronym{GAN}{GAN}{Generative Adversarial Network}
\newacronym{WGAN}{WGAN}{Wasserstein Generative Adversarial Network}
\newacronym{CycleGAN}{CycleGAN}{Cycle-Consistent Generative Adversarial Network}
\newacronym{ProGAN}{ProGAN}{Progressively-growing GAN}
\newacronym{ANN}{ANN}{Artificial Neural Network}
\newacronym{CNN}{CNN}{Convolutional Neural Network}
\newacronym{IoP}{IoP}{Internet of Production}
\newacronym{BNN}{BNN}{Bayesian Neural Network}
\newacronym{VIM}{VIM}{Vocabulaire International de Metrologie}
\newacronym{GUM}{GUM}{Guide to the uncertainty in measurement}
\newacronym{DMC}{DMC}{Data Matrix Code}
\newacronym{DL}{DL}{Deep Learning}
\newacronym{CE}{CE}{Circular Economy}
\newacronym{IR}{IR}{Infrared}
\newacronym{UV}{UV}{Ultraviolet}
\newacronym{GDL}{GDL}{Generative Deep Learning}
\newacronym{GenAI}{GenAI}{Generative Artificial Intelligence}
\newacronym{ID}{ID}{Inherent Dimension}
\newacronym{ML}{ML}{Machine Learning}
\newacronym{SEM}{SEM}{Scanning Electron Microscope}
\newacronym{AI}{AI}{Artificial Intelligence}
\newacronym{LG}{LG}{Logistic Regression}
\newacronym{SGD}{SGD}{Stochastic Gradient Descent}
\newacronym{PCA}{PCA}{Principal Component Analysis}
\newacronym{PC}{PC}{principal component}
\newacronym{SVM}{SVM}{Support Vector Machine}
\newacronym{LLM}{LLM}{Large-Language-Model}
\newacronym{EBM}{EBM}{Energy-based models}
\newacronym{DSR}{DSR}{Design Science Research}
\newacronym{MLP}{MLP}{Multi-Layer Perceptron}
\newacronym{GLCM}{GLCM}{Gray-level Co-occurence matrix}
\newacronym{CFRP}{CFRP}{Carbon-fiber reinforced plastics}
\newacronym{DLS-GAN}{DLS-GAN}{Defect Localization Sensitive \gls{GAN}}
\newacronym{StyleGAN}{StyleGAN}{Style-based Generative Adversarial Network}
\newacronym{DCGAN}{DCGAN}{Deep Convolutional Generative Adversarial Network}
\newacronym{ALAE}{ALAE}{Adversarial Latent Autoencoder}
\newacronym{AE}{AE}{Autoencoder}
\newacronym{AAE}{AAE}{Adversarial Autoencoder}
\newacronym{VAE}{VAE}{Variational Autoencoder}
\newacronym{HPO}{HPO}{Hyperparameter optimization}
\newacronym{JS}{JS}{Jenson-Shannon Distance}
\newacronym{FID}{FID}{Fr\'echet Inception Distance}
\newacronym{IS}{IS}{Inception Score}
\newacronym{SRAF}{SRAF}{Sub-resolution assist feature}
\newacronym{FD}{FD}{Fr\'echet Distance}
\newacronym{SeFa}{SeFa}{Semantic factorization}
\newacronym{MAE}{MAE}{Mean Absolute Error}
\newacronym{AQC}{AQC}{Automated Quality Control}
\newacronym[shortplural=MVS,firstplural=Machine Vision Systems (MVS), longplural=Machine Vision Systems]{MVS}{MVS}{Machine Vision System}
\newacronym{RMS}{RMS}{Root Mean Square}
\newacronym{DR}{DR}{Design requirement}
\newacronym{DP}{DP}{Design principle}
\newacronym{DF}{DF}{Design feature}
\newacronym{RMSE}{RMSE}{Root Mean Square Error}
\newacronym{ROI}{ROI}{Region of Interest}
\newacronym{SI}{SI}{International System of Units}
\newacronym{AdaIN}{\textsc{AdaIN}}{Adaptive Instance Normalization}
\newacronym[shortplural=CPPS]{CPPS}{CPPS}{Cyber-physical Production System}
\newacronym{RNN}{RNN}{Recurrent Neural Network}
\newacronym{PRISMA}{PRISMA}{Preferred Reporting Items for Systematic Reviews and Meta-Analyses}
\begin{document}

\title[Generative AI in Industrial Machine Vision - A Review]{Generative AI in Industrial Machine Vision - A Review}
\author*[1]{\fnm{Hans Aoyang} \sur{Zhou}\orcidlink{0000-0002-7768-4303}}\email{hans.zhou@wzl-iqs.rwth-aachen.de}
\equalcont{These authors contributed equally to this work.}

\author[1]{\fnm{Dominik} \sur{Wolfschl\"ager}\orcidlink{0000-0003-2399-4856}}\email{dominik.wolfschlaeger@wzl-iqs.rwth-aachen.de}
\equalcont{These authors contributed equally to this work.}

\author[1]{\fnm{Constantinos} \sur{Florides}\orcidlink{0000-0002-3031-9739}}\email{constantinos.florides@wzl-iqs.rwth-aachen.de}
\equalcont{These authors contributed equally to this work.}

\author[1]{\fnm{Jonas} \sur{Werheid}\orcidlink{0009-0003-6022-2633}}\email{jonas.werheid@wzl-iqs.rwth-aachen.de}
\equalcont{These authors contributed equally to this work.}

\author[1]{\fnm{Hannes} \sur{Behnen}\orcidlink{0009-0005-9369-8165}}\email{hannes.behnen@wzl-iqs.rwth-aachen.de}
\equalcont{These authors contributed equally to this work.}

\author[1]{\fnm{Jan-Henrik} \sur{Woltersmann}\orcidlink{0000-0003-0361-5302}}\email{jan-henrik.woltersmann@wzl-iqs.rwth-aachen.de}
\equalcont{These authors contributed equally to this work.}

\author[3]{\fnm{Tiago C.} \sur{Pinto}\orcidlink{0000-0002-8856-0648}}\email{tiago.pinto@ufsc.br}
\equalcont{These authors contributed equally to this work.}

\author[1]{\fnm{Marco} \sur{Kemmerling}\orcidlink{0000-0003-0141-2050}}\email{marco.kemmerling@wzl-iqs.rwth-aachen.de}

\author[1]{\fnm{Anas} \sur{Abdelrazeq}\orcidlink{0000-0002-8450-2889}}\email{anas.abdelrazeq@wzl-iqs.rwth-aachen.de}

\author[1,2]{\fnm{Robert H.} \sur{Schmitt}\orcidlink{0000-0002-0011-5962}}\email{robert.schmitt@wzl-iqs.rwth-aachen.de}

\affil*[1]{\orgdiv{Laboratory for Machine Tools and Production Engineering, WZL}, \orgname{| RWTH Aachen University}, \orgaddress{\street{Campus-Boulevard 30}, \city{Aachen}, \postcode{52074}, \country{Germany}}}

\affil[2]{\orgdiv{Fraunhofer IPT}, \orgaddress{\street{Steinbachstr. 17}, \city{Aachen}, \postcode{52074}, \country{Germany}}}

\affil[3]{\orgdiv{LABMETRO, EMC, Universidade Federal de Santa Catarina}, \orgaddress{\street{CP5053, 88040-970}, \city{Florianopolis}, \country{Brazil}}}

\abstract{

Machine vision enhances automation, quality control, and operational efficiency in industrial applications by enabling machines to interpret and act on visual data. 
While traditional computer vision algorithms and approaches remain widely utilized, machine learning has become pivotal in current research activities. 
In particular, generative \gls*{AI} demonstrates promising potential by improving pattern recognition capabilities, through data augmentation, increasing image resolution, and identifying anomalies for quality control.
However, the application of generative \gls*{AI} in machine vision is still in its early stages due to challenges in data diversity, computational requirements, and the necessity for robust validation methods.
A comprehensive literature review is essential to understand the current state of generative \gls*{AI} in industrial machine vision, focusing on recent advancements, applications, and research trends. Thus, a literature review based on the PRISMA guidelines was conducted, analyzing over 1,200 papers on generative \gls*{AI} in industrial machine vision. 
Our findings reveal various patterns in current research, with the primary use of generative \gls*{AI} being data augmentation, for machine vision tasks such as classification and object detection. Furthermore, we gather a collection of application challenges together with data requirements to enable a successful application of generative \gls*{AI} in industrial machine vision. This overview aims to provide researchers with insights into the different areas and applications within current research, highlighting significant advancements and identifying opportunities for future work.
}

\keywords{Machine Vision, Generative Artificial Intelligence, Deep Learning, Machine Learning, Manufacturing}
\maketitle              

\section{Introduction}
\label{sec:introduction}

Visual inspection performed by trained inspectors is still widely used in industry, but since the 1970s, automated machine vision has been systematically introduced \cite{Beyerer.2016}. 
Industrial machine vision, an essential component of modern manufacturing processes, involves the processing and analysis of images to automate tasks, including quality inspection, object or defect detection, and process control \cite{Smith.2021}. Traditional computer vision systems rely on classical algorithms and techniques, that require hand-crafted features, which, although practical, have limitations in handling complex scenarios with significant variability and unforeseen cases \cite{Bhatt.2021, Smith.2021}. In the 1980s and 1990s, technology advanced with techniques such as digital image processing, texture, and color analysis, supported by better hardware and software \cite{Manakitsa.2024}. It relied on predefined algorithms for tasks like quality inspection and object recognition \cite{Bhatt.2021, Chin.1982}.

The late 1990s and early 2000s saw a shift towards machine learning, where models like \glspl*{SVM} \cite{Hearst.1998}, Random Forests \cite{Breiman.2001}, and \glspl*{ANN} enabled systems to learn in a data-driven way, improving their performance to handle real-world variability and complexity \cite{Smith.2021}.
The true revolution in machine vision came along with the development of \gls*{DL} in the 2010s. \glspl*{CNN} have proved exceptionally powerful for image processing tasks. \glspl*{CNN} enabled machines to automatically learn hierarchical features from raw image data \cite{Cao.2021}, vastly improving performance on tasks such as image classification, image segmentation, object detection, defect detection, and pose estimation \cite{Chen.2021, Jha.2023, Shavit.08.07.2019, Manakitsa.2024}. Landmark models like AlexNet, VGG, and ResNet showcased the potential of \gls*{DL}, leading to rapid adoption in both academic research and industry \cite{Smith.2021}.

\gls*{GenAI} represents the latest frontier in the evolution of machine vision. Unlike traditional discriminative models that classify or recognize patterns, \gls*{GenAI} models can create new data instances. While most popular \gls*{GenAI} models and innovations are designed for human interaction, there is a significant opportunity to explore how \gls*{GenAI} can transform industrial manufacturing. 
Comparable alternatives for data generation like simulations require expert domain knowledge and manual execution. 
Thus, for industrial manufacturing applications, their use is limited to the pre-processing and post-processing steps. 
Whereas, \gls*{GenAI} methods once trained have the potential to automate currently manual processing steps during manufacturing.
Due to its promising potential, \gls*{GenAI} has been applied to different machine vision use cases, where each proposed solution was developed under its use case specific constraints.
This collection of findings and experiences compiled over the machine vision research landscape hold valuable insights for other practitioners that aim to use \gls*{GenAI} for their own research purposes.   
Despite the existing knowledge of applying \gls*{GenAI} in various machine vision use cases, to the best of our knowledge, there is no review dedicated to \gls*{GenAI} in the context of industrial machine vision that consolidates the available application experience. 
The only literature reviews that mention \gls*{GenAI} within the context of industrial machine vision, focus on \gls*{AI} in general applied to industrial machine vision tasks within specific manufacturing domains like printed circuit boards \cite{Andrade.2022}, silicon wafers \cite{Batool.2021}, general defect recognition \cite{Gao.2022}, or surface defect recognition \cite{Prunella.2023}. 

This reviews contributions are: (i) it gives a general overview about \gls*{GenAI} methods used in industrial machine vision applications, (ii) provides an overview of the tools, potentials, and challenges when applying \gls*{GenAI}, and (iii) presents the benefits of \gls*{GenAI} in typical machine vision applications for practitioners.

From the objectives, we derive the following research questions addressed in this review:
\begin{enumerate}
\item Which \gls*{GenAI} model architectures are used within industrial machine vision applications?
\item Which requirements and properties must \gls*{GenAI} methods fulfill to be transferable to the domain of industrial machine vision?
\item To which industrial machine vision tasks have \gls*{GenAI} successfully been applied?
\end{enumerate}

This work is structured as follows. First, an overview of the field and methods of \gls*{GenAI} is given in Section~\ref{sec:generative_ai}. Section~\ref{sec:methodology} presents the methodology used for conducting the literature review, including a comprehensive justification of the derivation of exclusion criteria and the choice of the information to be extracted from the literature. Section~\ref{sec:literature_review} presents the search results and its characteristics, followed by an extensive analysis of the extracted data. The results of the literature review are discussed with respect to the research questions in Section~\ref{sec:discussion}. The discussion also concludes with a reflection of the biases and limitations of the applied literature review methodology. The paper concludes, by outlining the central results of the review and pointing out guidelines for the application of \gls*{GenAI} in the industrial machine vision tasks.

\section{Generative Artificial Intelligence}
\label{sec:generative_ai}
The field of \gls*{GenAI} represents semi-supervised and unsupervised \gls*{DL} techniques that aim to learn the probability distribution $p\li x \re$ of a given dataset $x \in \mathcal{X}$. In the context of \gls*{DL}, \gls*{GenAI} methods approximate the probability distribution $p\li x \re$ using \glspl*{ANN} that are parameterized with weights $\Theta$, resulting in a parametric model $p_{\Theta} \li x \re$. Compared to discriminative \gls*{DL} techniques, which approximate a probability distribution $p\li y | x\re$ over an attribute (or label) $y$ given an input $x$, generative models $\mathcal{G}$ can be used to draw samples $\tilde{x} \sim p_\Theta \li \tilde{x} \re$ that resemble instances from the training data distribution~\cite{BondTaylor.2022}.

\begin{figure}[htb]
    \centering
    \includegraphics[width=\textwidth]{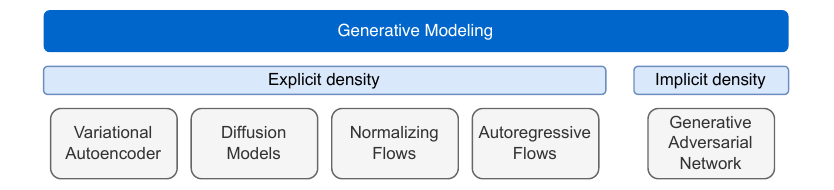}
    \caption[Taxonomy of GenAI approaches.]{\label{fig:generative_deep_learning_taxonomy}Taxonomy of GenAI approaches. The task of density estimation can be achieved through an explicit or implicit density estimation. Adapted from~\cite{Foster.2023}.} 
\end{figure}

The estimation of $p\li x \re$ can be divided into \textit{explicit} and \textit{implicit} approaches. While explicit estimation models try to provide a parametrization of the probability density $p_{\Theta} \li x \re$, implicit estimation models build a stochastic process that synthesizes data~\cite{Foster.2023}. 
An overview about the taxonomy of \gls*{GenAI} (cf. \figreft{fig:generative_deep_learning_taxonomy}) summarizes existing approaches to estimate $p_{\Theta} \li x \re$. Independent of the model type, their ability to generate photorealistic high-resolution images has attracted their use in solving classical computer vision tasks like image inpainting, image denoising, image-to-image translation, and other image editing problems. Their promising performance in academic benchmarks, make them relevant for the domain of machine vision. Further descriptions of each model architecture with their advantages and constraints will be explored in the following subsections.

\subsection{Variational Autoencoders}
\begin{figure}[htb]
    \centering
    \includegraphics[width=\textwidth]{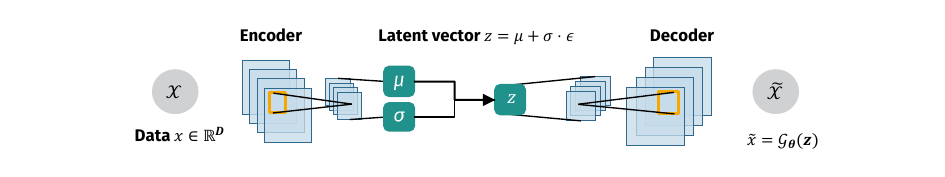}
    \caption{The \gls*{VAE} architecture is displayed with both encoder and decoder, where the encoder encodes the input data $x$ with dimensions $D$ into a representation of mean $\mu$ and standard deviation $\sigma$ values, resulting together with $\epsilon \sim \scriptN\li 0,1 \re$ in the latent variable $z = \mu + \sigma \epsilon$. Afterward, the decoder $\mathcal{G}$ decodes the latent variable back into an image $\tilde{x} = \mathcal{G}_{\Theta} \li z \re$, with weights $\Theta$.}
    \label{fig:VAE}
\end{figure}

Derived from the assumption that images are generated from an unknown source described by a latent vector $z$, \textit{Latent Variable Models} $p_\Theta \li x | z \re$ use \glspl*{CNN} to generate samples $x$ from a prior distribution $p\li z\re$.
One of the most prominent latent variable models is the \gls*{VAE} proposed by \textsc{Kingma}~\cite{Kingma.2014}, which extends a deterministic autoencoder architecture with a probabilistic latent variable model $p_\Theta \li x, z\re=p_\Theta \li x |z\re p_\Theta \li z\re $.
As shown in Figure \ref{fig:VAE}, \glspl*{VAE} consist of an encoder, a decoder, and the probabilistic latent variable.
The \gls*{VAE} solves the challenge of the unregularized distribution of the latent space of autoencoder models by imposing a (multi-variate) normal distribution $\scriptN$ into the latent space: 
\begin{equation}
    \scriptN\li x; \mu, \sigma\re = \frac{1}{\sqrt{2\pi\sigma^2}}e^{-\frac{\li x-\mu \re^2}{2\sigma^2}}
\end{equation}
However, due to the Gaussian prior, \glspl*{VAE} will likely generate blurry images on larger resolutions stems \cite{Dosovitskiy.2016}. Nonetheless, the \gls*{VAE} is a fundamental and well-known architecture within the field of \gls*{GenAI}.

\subsection{Diffusion Models}
\textit{Diffusion models}~\cite{SohlDickstein.2015, Croitoru.2023} represent a class of probabilistic models that use a Markov process of $T$ steps to gradually transform a sample $x_0$ into noise. At each time step $t$, this forward noising process applies Gaussian noise with variance $\beta_t$ to the sample and can be described formally with the identity matrix $I$ as follows 
\begin{equation}
    x_T \sim \scriptN: p\li x_t | x_{t-1} \re = \scriptN \li x_t; \sqrt{1-\beta_t}x_{t-1}, \beta_t \cdot I  \re.
\end{equation} 
In the reverse direction, a generative model is trained to remove the noise added in one step $t$.
They generate images from pure noise by applying the reverse diffusion process for $T$ steps. Since 2020, the popularity of diffusion models has increased significantly, particularly when conditional diffusion processes through combination with models like CLIP \cite{Radford.2021} were introduced. Noteworthy contributions, such as Stable Diffusion~\cite{Rombach.2022}, and OpenAI's DALL-E~\cite{Ramesh.2022}, have additionally played pivotal roles in elevating the recognition of diffusion models. 
However, despite the stable training and the high diversity in the generation process, the necessity of applying the model $T$ times to generate one image poses the challenge of low inference speed. 

\subsection{Normalizing Flows}
\textit{Normalizing flows} are identical to \gls*{VAE} in the sense, that they are able to encode a complex distribution (like an image) into a simple distribution (normal distribution) and vice versa. However, \glspl*{VAE} encoder and decoder are different models with different weights, whereas for normalizing flows, the encoding and decoding is done with the same invertible model $p_\Theta \li x \re$. This invertible model consists of a sequence of invertible functions $f: \mathbb{R}^d \rightarrow \mathbb{R}^d$ with their corresponding inverse function $g=f^{-1}$. When applying this function to a random variable $x \sim p\li x \re$, the distribution of the resulting random variable $y = f\li x \re$ is yield by the change of variables rule: 
\begin{equation}
    p\li y \re = p\li x \re \left| \text{det} \partialderivative{f^{-1}}{y} \right| = p\li x \re \left| \text{det} \partialderivative{f}{y}\right|^{-1} 
\end{equation} 
Given a series of $K$ inverse mapping functions, a variable with distribution $p_0$ can be transformed into an arbitrarily complex density $p_K\li x_K\re$ with 

\begin{equation}
    \begin{aligned}
        x_K &= f_K \circ \ldots  \circ f_2 \circ f_1\li x_0 \re \\
        \ln p_K \li x_K\re &= \ln p_0 \li x_0\re - \sum_{k=1}{K} \ln \left|\text{det} \partialderivative{f_k}{x_{k-1}}\right|
    \end{aligned}    
\end{equation}

Normalizing flows provide flexibility when it comes to generative modelling because of their precise likelihood evaluation and efficient sampling. However, for large datasets, there is a trade-off since a larger amount of training data means an increase in the need for computational resources \cite{Bandi.2023}.

\subsection{Autoregressive Models}

\textit{Autoregressive models} utilize the chain rule of probability to break down the joint probability of a set of variables into a sequence of conditional probabilities. Mathematically, this is expressed as 

\begin{equation}
     p(x) = p(x_1, x_2, \ldots, x_n) = \prod_{i=1}^{n} p(x_i \mid x_1, x_2, \ldots, x_{i-1}). \ 
\end{equation}

Based on each sequence of conditional probabilities, Autoregressive Models can directly maximize the likelihood of predicting data by minimizing the negative log-likelihood. However, increasing the dimensionality of provided data, such as images, negatively impacts their sampling time. Furthermore, data needs to be broken down into specific orders. The selection order is evident for certain modalities, like text and audio. However, for other modalities, such as images, this is not the case, and it can affect the performance of the network architecture in use \cite{BondTaylor.2022, Turhan.2018}.

To address these challenges, various architectural improvements were introduced, such as in Masked \glspl*{MLP} using time-dependent masks to ensure the autoregressive property or in \glspl*{RNN} which are suitable for sequential data modeling \cite{BondTaylor.2022}. Also, autoregressive models are applied for analyzing images pixel-by-pixel and are combined with various neural network architectures such as \gls*{CNN}, \gls*{RNN}, \gls*{VAE}, etc. to improve their modeling performance \cite{Turhan.2018}. However, due to the sequential sampling method of autoregressive models, their sampling time is usually too high for use cases with real-time constraints.  

\subsection{Generative Adversarial Nets}

\begin{figure}[htb]
    \centering
    \includegraphics[width=\textwidth]{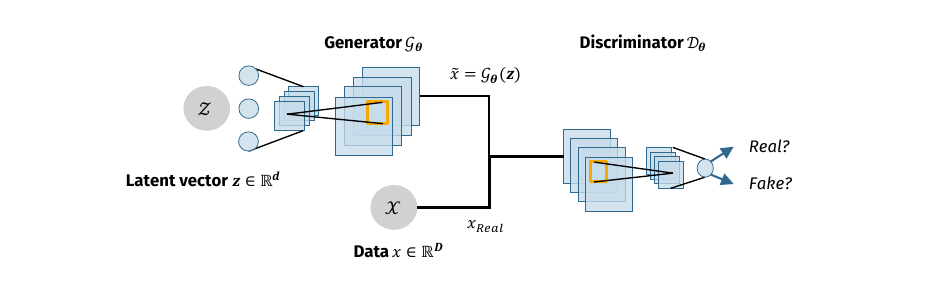}
    \caption[Structure of a vanilla \gls*{GAN}.]{The \gls*{GAN} architecture is displayed with both generator $\mathcal{G}_\Theta$ and discriminator $\mathcal{D}_\Theta$, both parameterized with weights $\Theta$. During training, $\mathcal{G}_\Theta$ generates a fake image $\tilde{x} \in \mathbb{R}^D$ from a latent vector $z \in \mathbb{R}^d$. Afterward, $\tilde{x}$ and $x_{Real}$ are both used to train $\mathcal{D}_\Theta$, which tries to predict whether the image is from the real data distribution $p\li x \re$ or the fake data distribution $p_{\mathcal{G}}\li x \re$.} 
    \label{fig:GAN}
\end{figure}

The recent success of \gls*{GenAI} is founded on the development of the \gls*{GAN} architecture as depicted in figure \ref{fig:GAN}. \glspl*{GAN}, first proposed by \textsc{Goodfellow} in 2014, use a technique from game theory to train a conglomerate of a generator network $\mathcal{G}$ and a discriminator network $\mathcal{D}$ ~\cite{Goodfellow.2014}. The generator represents a mapping function $\mathcal{G}: \mathbb{R}^d\rightarrow \mathbb{R}^D$  that takes a $d$-dimensional vector $z\sim p\li z\re$ sampled from a simple prior distribution as input to generate a synthetic (fake) image $\tilde{x} \in \mathbb{R}^D$ according to the learned distribution $p_\mathcal{G} \li x \re$. Usually, the discriminator represents a function $\mathcal{D}: \mathbb{R}^D\rightarrow \left[0,1\right]$ which predicts whether a given real image $x$ or synthetic image $\tilde{x}$ belongs to the data distribution $p\li x \re$. In this way, the challenge of providing an objective function that allows to optimize the generator parameters to sample from the data distribution is reformulated by means of a binary classification task. The objective function for training the two networks is~\cite{Goodfellow.2014}:
\begin{equation}
    \mathcal{L}_\text{Vanilla GAN} = E_{ x \sim p \li x \re } \Bigr[ \text{log}\bigl(\mathcal{D} \li x\re \bigl) \Bigr]+ E_{z\sim p  \li z\re} \biggr[ \text{log} \Bigl( 1-\mathcal{D} \bigl( \mathcal{G} \li z\re \bigl) \Bigl) \biggr]
\end{equation}
Thereby, the difference between $p(x)$ and $p_\mathcal{G} \li x \re$ is measured by the discriminator and used to refine the weights of the generator to generate samples that resemble those from the data distribution $p(x)$. During training, the discriminator is exposed to alternating synthetic images from $\mathcal{G}$ and real images to effectively learn to classify real and fake images. The generator uses the feedback of the discriminator to learn to produce more realistic synthetic images to deceive the discriminator and ultimately approximate the intractable true probability distribution $p_\mathcal{G} \li x \re \approx p(x)$. This is called an adversarial game, because the generator tries to maximize the probability to deceive the discriminator and the discriminator follows the opposite goal. The min-max optimization problem tries to find the Nash equilibrium, which corresponds to finding a saddle point in the landscape of $\mathcal{L}_\text{Vanilla GAN}$. This makes the training of \glspl*{GAN} particularly instable. One possible issue is the \textit{mode collapse} phenomenon, which occurs when one of the networks learns too fast while the other cannot catch up with it, so that the feedback gradient vanishes~\cite{Bengesi.2023}.

Numerous enhanced \gls*{GAN} model architectures were developed since then. In the vanilla GAN implementation~\cite{Goodfellow.2014}, $\mathcal{D}$ and $\mathcal{G}$ are composed of feed-forward neural networks and therefore capable of generating realistically-appearing images only at small resolutions. To improve the quality of generated images, the \gls*{DCGAN} introduced the usage of deep \glspl*{CNN}. In their work, \textsc{Radford} also noticed that the latent space of \gls*{DCGAN} allows for composing visual semantics using vector arithmetic in the latent space~\cite{Radford.2015}. Moreover, the development of \gls*{WGAN} increased the training stability of \glspl*{GAN} by introducing a smoother gradient for the generator using the \textit{Earth-Mover} or \textit{Wasserstein} distance instead of the cross-entropy during training~\cite{Arjovsky.2017}.

\begin{figure}[ht!]
    \centering
    \includegraphics[width=\textwidth]{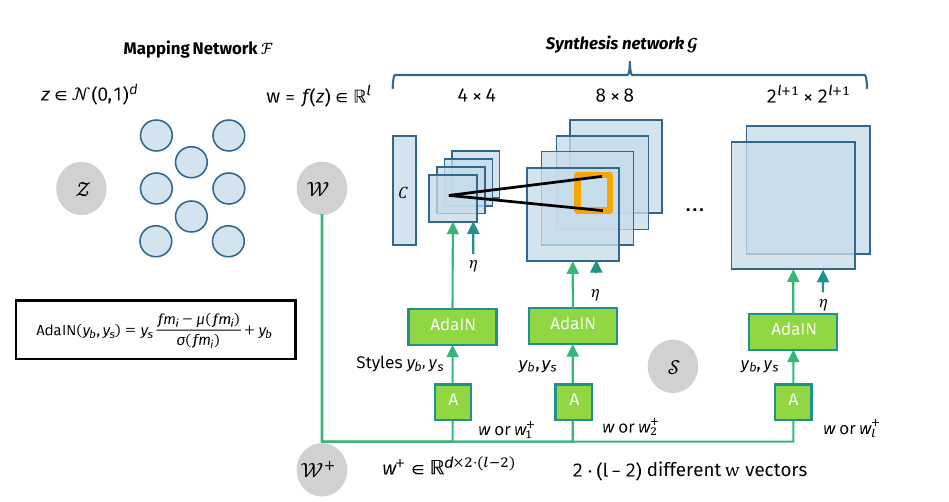}
    \caption[Architecture of the \gls*{StyleGAN} model.]{\label{fig:StyleGAN}Simplified architecture of the \gls*{StyleGAN} model showing the mapping network $\mathcal{F}$ and the progressively growing synthesis network $\mathcal{G}$ with the different latent spaces.} 
\end{figure}

A breakthrough for the generation of high-resolution images was achieved by the development of \gls*{ProGAN}~\cite{Karras.2017}, that made it possible to synthesize high-resolution images up to $1024\times 1024\unit{pixels}$. Progressive growing refers to a training strategy where the resolution of training images is gradually increased, as indicated in \figreft{fig:StyleGAN}. This, improves the training stability of \glspl*{GAN} at higher resolutions, because at lower resolution ($4\times 4\unit{pixels}$) learning visual concepts that can compete with the discriminator is of lower complexity for the generator. In early training epochs only the first layer is trained, in the end the network is trained at full resolution with all $l$ generator layers ($2^{l+1}\times 2^{l+1}\unit{pixels}$). This allows the generator to keep track with the discriminator by gradually increasing complexity and the level of detail with the resolution. 

The \gls*{StyleGAN} architecture extends the \gls*{ProGAN} architecture by introducing an intermediate latent space $\latW$ and the concept of neural style transfer~\cite{Karras.2021}. $\latW$ is mapped from the normal prior $\latZ$ using a Fully Connected Network, called mapping network $\mathcal{F}$. This allows $\latW$ to form freely during training and approximate an advantageous and natural distribution. The intermediate latent space vectors $w\in \latW$ are injected into the progressively growing layers of the synthesis network  $\mathcal{G}$, which allows for controlling image properties at different resolution levels. Thereby, the generation process applies \textit{styles} and additional noise vectors $\eta$ at different resolution levels to a learned constant $C$, which represents the center of $p_G \li x \re$. The architecture of the \gls*{StyleGAN} model is depicted in Figure \ref{fig:StyleGAN}. For image generation, a vector is sampled in $\latZ$ and transformed into an intermediate latent space vector $w \in \latW$. This vector is fed into the individual layers of the synthesis network $\mathcal{G}$ (thereby spanning up the so-called extended intermediate latent space $\latW^+$) and transformed, using an affine transformation $A$, into a style bias $y_{b}$ and scaling vector $y_{s}$. Afterwards, Adaptive Instance Normalization (\textsc{AdaIN})

\begin{equation}
    \label{eq:adain}
    \textsc{AdaIN}(G_l, y_s, y_b) = y_{s}\frac{G_l-\mu(G_l)}{\sigma(G_l)}+y_{b},
\end{equation}

is applied on every synthesis feature map $G_l$, where $l$ describes the number of convolutional layers in $\mathcal{G}$ ~\cite{Huang.2017}. In this way, the constant $C$ is modulated with the learned \textit{styles}, where each injected latent vector controls a specific feature map at the corresponding resolution.
Compared to other \gls*{GAN} models, the resulting latent spaces of \gls*{StyleGAN} models are smoother and disentangled; Each dimension corresponds to an individual semantic property of the synthesized image. Once each dimension has been interpreted, \glspl*{StyleGAN} can be used to freely adjust the image generation process. \glspl*{StyleGAN} and their variants represent the current state of the art \gls*{GAN}-based image synthesis model architecture with respect to resolution, image quality and control over generated features~\cite{Karras.2021}.

Multiple upgrades of the \gls*{StyleGAN} architecture have been presented to address limitations and artifacts of the initial architecture design. Particularly for \gls*{StyleGAN}\,2  the progressive growing strategy is replaced by a skip-connection-based architecture, such that the generator is created by summing up residuals of each resolution block~\cite{Karras.2019}. On one hand this decreased the frequency of blob-artifacts in the synthesized images, on the other hand it made it possible to embed images into the latent space of \gls*{StyleGAN}. Further improvements proposed in the \gls*{StyleGAN}\,3 architecture concentrated on resolving the problem of \textit{texture-sticking} through various small architectural edits. However, the changes introduced new artifacts and apparently lead to a lower degree of disentanglement of learned representations~\cite{Alaluf.2022}. 


\section{Research Methodology}
\label{sec:methodology}
As stated in the \nameref{sec:introduction}, this literature review aims to provide an overview of \gls*{GenAI} methods and applications within the field of industrial machine vision for manufacturing applications.
It was conducted based on the \gls*{PRISMA} method, which is designed for presenting and generating systematic reviews in a transparent, complete, and accurate manner \cite{MatthewJ.Page.2021}. Given this method, the following sections present the approach of the systematic review. 
Initially, eligibility measure in the form of exclusion criteria are introduced together with the search strategy as well as the utilized literature databases (cf. Section \ref{subsec:search_strategy}). Followed by the remaining two sections, the study selection process (cf. Section \ref{subsec:study_selection}) and data extraction (cf. Section \ref{subsec:data_extraction}). 

\subsection{Search Strategy and Databases}
\label{subsec:search_strategy}
To identify relevant literature, exclusion criteria were constructed (cf. Table \ref{tab:exclusion_criteria}), that build the foundation for the selection process of all retrieved documents during abstract screening and full-text reviews. These criteria ensure that only publications are identified, that are relevant for the research scope defined by the research questions. 
For the extraction of literature, the databases \textit{Scopus}, \textit{Web of Science} and \textit{IEEE Xplore} are used. 
They cover a wide range of different topics from engineering to computer science with a balanced mix of conference proceedings and journal publications. 

\begin{table}[htbp]
\centering
\caption{Exclusion criteria for selecting relevant literature}
\label{tab:exclusion_criteria}
\begin{tabular}{p{.12\textwidth} p{.3\textwidth} p{.45\textwidth}}
\toprule
Criteria No. & Description & Reasoning \\
\midrule
1 & Published before 2018 & First applications of \gls*{GenAI} beyond academic developments which demonstrated high sampling quality for large resolutions were shown in 2018. \\
2 & Not in English & Providing an overview of English written literature enables traceability for most readers. \\
3 & Sole application of discriminative models & Applied AI methods used should contain at least to some extent generative models. \\
4 & Image generation without AI & Simulation software or standard data augmentation methods are also capable of generating new data instances, but are defined as out of scope. \\
5 & \gls*{GenAI} for other modalities than images & Only vision-based \gls*{GenAI} is analyzed within this review, because other modalities are not directly applicable to industrial machine vision tasks. \\
6 & Not related to an industrial domain & Main contribution of review is the focus on industrial applications. Application areas not relevant for industrial purposes are therefore excluded.\\
\bottomrule
\end{tabular}
\end{table}

After selecting eligibility criteria and deciding on information sources, the next step in the \gls*{PRISMA} methodology is to define the search strategy. This includes the construction of a search string, where each keyword was selected based on an iterative exploratory analysis;
Different keyword combinations were evaluated based on their estimated ratio of relevant publications. 
The resulting search string for this review as follows: 

\begin{displayquote}
\texttt{((Generat* OR GAN OR Diffusion Model OR Normalizing Flow OR Autoencoder) AND (Artificial* OR Machine Learning OR Deep Learning OR Neural Network) AND (Industr* OR Manufact* OR Production*) AND (Image* OR Vision OR Optical OR Visual*) AND (Quality OR Metrology OR Monitoring))}.
\end{displayquote}

Similar to the already mentioned exclusion criteria 1 and 2, the search string was complemented with the following filters to further narrow down the identified documents:

\begin{itemize}
    \item \textbf{Language.} Only English literature was chosen.
    \item \textbf{Year.} Only literature published after 2018 was categorized as relevant.
    \item \textbf{Research Area.} Documents from the domains of engineering and material science were selected to be relevant.
\end{itemize}

With the search string defined, and the search parameters configured, the search was executed in September 2023 using the databases listed before. Publications in September 2023 and later are not included within this review.


\subsection{Study Selection} 
\label{subsec:study_selection}
\begin{figure}
    \centering
    \includegraphics[width=1\linewidth]{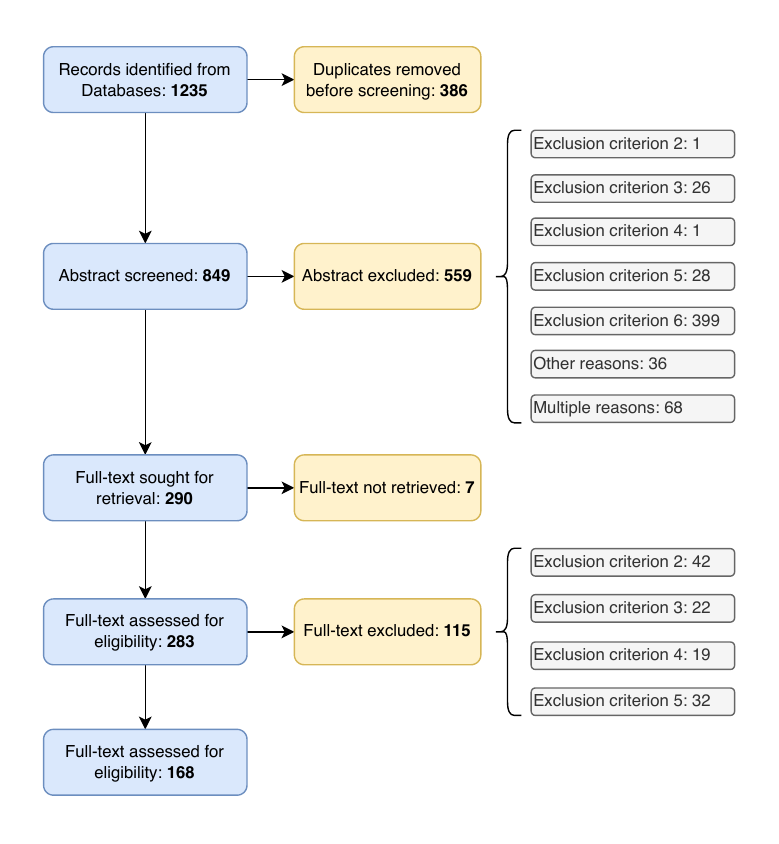}
    \caption{\gls*{PRISMA} flowchart showing the number of publications excluded during study selection.}
    \label{fig:prisma-flowchart}
\end{figure}

For study selection, a two-step process was applied, starting with an abstract screening to filter out the vast majority of irrelevant publications, followed by a full-text review. To ensure a high review quality, a dual-review with a principle reviewer was used for the abstract screening stage. That is, each abstract was screened by two reviewers, and in case of different opinions a third reviewer was conducted for a final decision. For both review rounds, we use the previously defined exclusion criteria and kept publications with no clear exclusion criterion during abstract screening for full-text review. After title and abstract screening, during full-text reviews, we analyzed whether the scope of the publication still fits our eligibility criteria. This review process was designed as a single review process, so that each full-text publication was evaluated by one reviewer.
An overview about the study selection process, where the number of publications removed at each stage as well as their reason for removal, is shown in the \gls*{PRISMA} flowchart depicted in Figure \ref{fig:prisma-flowchart}.

As Figure \ref{fig:prisma-flowchart} shows, only 168 papers from an initial, 1235 retrieved documents persisted to the filtering stages. $386$ publications were sorted out due to duplicates. A majority of $399$ articles were excluded because they were not relevant for any industrial use case. Oftentimes the relevance for industrial use cases is claimed, without demonstrating \gls*{GenAI} to an industrial use case. The high number of records excluded for other reasons consist mainly of review papers from different domains, but not directly addressing any research question from this review.

\subsection{Data Extraction}  
\label{subsec:data_extraction}

All publications that successfully passed the full-text screening underwent data extraction. Its purpose is to extract relevant information from each publication to answer the previously defined research questions.
For the data extraction process, a list of predefined categories, each representing pertinent content relevant to a research question.
For answering the first research question, the model architectures were investigated to extract an overview about the distribution of applied model architectures. Initially the exact model architecture was listed, and afterward they were grouped into model families and their derivatives. For the second research question, general success factors for applying \gls*{GenAI} methods were investigated during data extraction. By analyzing dataset and model architecture properties, the goal was to extract repeating patterns for successfully applying \gls*{GenAI} methods. Finally, for answering the third research question, the machine vision tasks together with the \gls*{GenAI} purpose were collected, investigating the use of \gls*{GenAI} for different machine vision tasks. 
Consequently, besides a basic summary of the contribution for each paper, the following categories were extracted from each publication:
\begin{itemize}
    \item \textbf{Model architecture.} Which model family (e.g., GAN, VAE etc.) and which architecture (e.g., StyleGAN, CycleGAN etc.) are used?
    \item \textbf{Dataset information.} What dataset is used, and what are its properties (e.g., number of entities, resolution etc.)?
    \item \textbf{Properties of \gls*{GenAI} model.} Based on which properties of a specific architecture is the generative model selected?
    \item \textbf{Data requirements.} What requirements and limitations were mentioned with respect to the training dataset?
    \item \textbf{Machine vision task.} For which specific machine vision task was the \gls*{GenAI} model utilized (e.g., classification, segmentation etc.)
    \item \textbf{Purpose of \gls*{GenAI}.} For what reason are \gls*{GenAI} models applied to the machine vision task (e.g., data augmentation, image reconstruction etc.)
\end{itemize}

The categories are initially filled manually by one reviewer, detailing relevant information from each publication. Subsequently, patterns were searched and, if possible, placed into discrete clusters. This clustering simplifies the subsequent analysis of quantitative information. After extracting data from defined categories and organizing them into quantitative clusters, important correlations were analyzed. These results will be presented within the next section.

\section{Literature Analysis}
\label{sec:literature_review}
With the literature review process defined, the results of the review are presented in the following. According to the research questions defined in Section \ref{sec:introduction}, each section aims to answer a research question. First, an overview about \gls*{GenAI} architectures in machine vision applications is presented in Section \ref{subsec:gen_ai_in_machine_vision}. Next, challenges and requirements for \gls*{GenAI} are presented in Section \ref{subsec:application_challenges}. Finally, the application of \gls*{GenAI} for various industrial machine vision tasks is analyzed in Section \ref{subsec:machine_vision_tasks}.

\subsection{Generative Artificial Intelligence Architectures used in Industrial Machine Vision}
\label{subsec:gen_ai_in_machine_vision}
\begin{figure}[ht!]
    \centering
    \includegraphics[width=.7\textwidth]{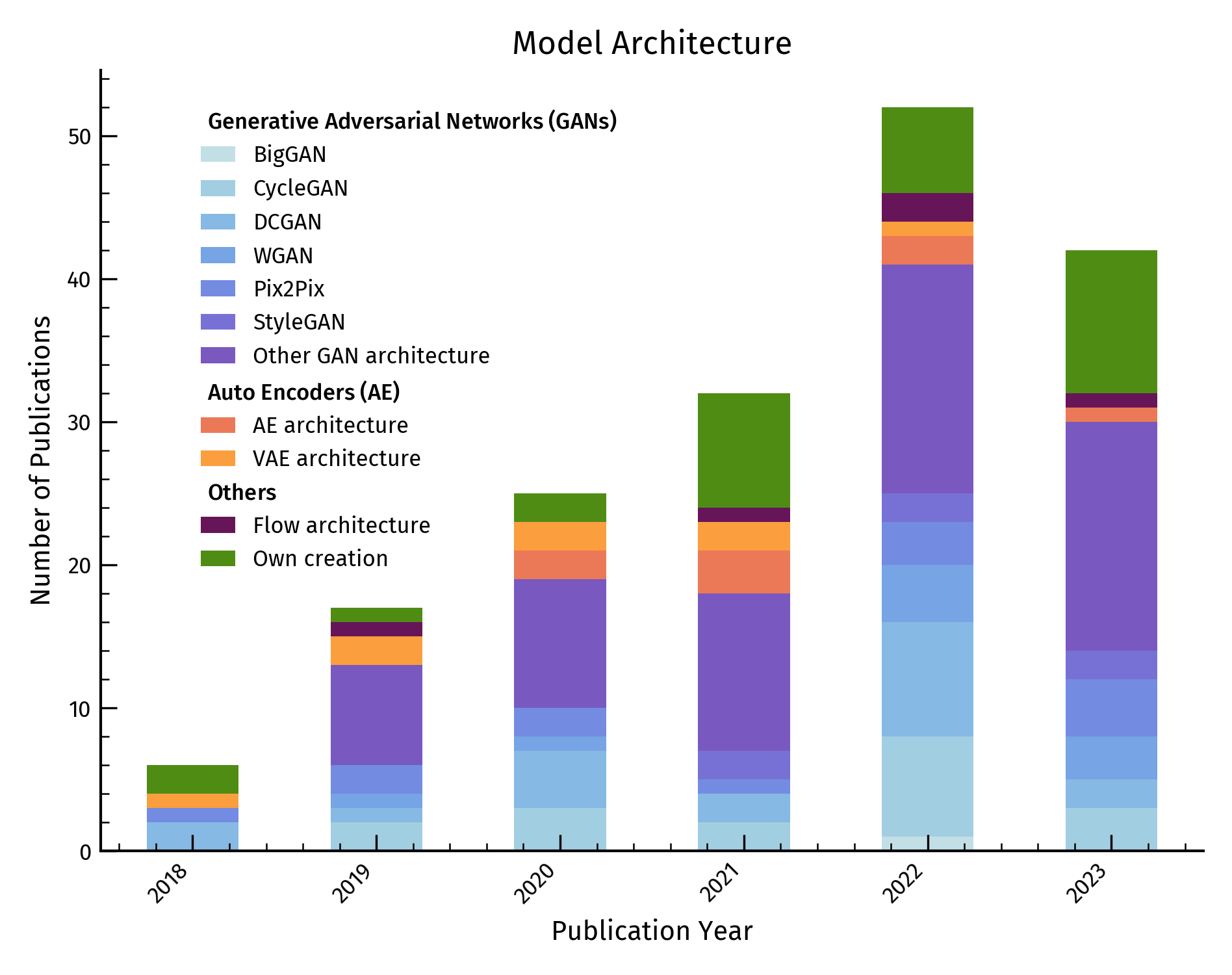} 
    \caption{Publication trends of \gls*{GenAI} technologies in industrial machine vision.}
    \label{fig:architecture_over_year}
\end{figure}

From Section \ref{sec:introduction}, it was already introduced that there is a rising interest for \gls*{GenAI} in industrial machine vision. In order to confirm this trend, a continuous increase in the total number of reviewed publications over the years could be observed, as shown in Figure \ref{fig:architecture_over_year}. 
Looking further into research question 1, the architecture distribution shows clearly that the majority of publications use \gls*{GAN}-based architectures, followed by \gls*{VAE}-based architectures. Only five publications use Flow-based architectures, and to the best of our knowledge no diffusion-based architectures or autoregressive architectures were used for industrial machine vision. The absence of these architectures may be due to the fact that the image sampling time is too high for industrial applications.   

\figreft{fig:architecture_over_year} further reveals that a significant amount (ca. $20 \%$) of publications customized the model architecture to fit the specific industrial machine vision use case. Most adjustments adapt an available architecture, such as the \gls*{DCGAN} architecture. Although \gls*{StyleGAN} demonstrate advantageous properties in sampling quality and manipulation, only seven publications applied them for their work. This either showcases an unexploited improvement potential for currently proposed \gls*{GenAI} uses or that the application of \gls*{StyleGAN} entails currently unsolvable challenges. Either way, these findings showcase further research is necessary to simplify the transfer of state-of-the-art \gls*{AI} results to manufacturing applications.


\subsection{Properties for Successfully Applying Generative Artificial Intelligence Models to Industrial Machine Vision}
\label{subsec:application_challenges}
The previously already demonstrated low number \gls*{StyleGAN} of applications in industrial machine vision indicate application challenges of \gls*{GenAI} architectures.
In order to answer the second research question regarding which requirements and properties must \gls*{GenAI} methods fulfill in order to be applicable within the domain of industrial machine vision, this review investigated the relevant properties of the model architecture with their encountered challenges and limitations. Furthermore, requirements of the data used for training \gls*{GenAI} models were also analyzed. 
From our data extraction process, the following general themes that determine the success of \gls*{GenAI} transferability were identified.

\paragraph{Evidence of Practical Use in Other Domains.}
The first reason for practitioners in industrial machine vision to use a particular \gls*{GenAI} model architecture is when evidence of practical applicability in other domains can be shown. It could be observed, that for example, the ability to solely sample realistic images from a learned probability distribution is not a sufficient reason to apply a generative model. However, through leveraging the generative modelling principles to practical applications, their transfer to industrial machine vision becomes more likely. Most common, these applications are image-to-image translation \cite{Holscher.2022, Alam.2023, Liu.2022b, Noraas.2019, Posilovic.2022, Hoq.2023, Wang.2021c, Baranwal.2019}, image-enhancement \cite{Zhang.2022, Eastwood.2021, Eastwood.2022}, feature extraction capability \cite{Tan.2023, Tang.2023, Zhang.2024, Lu.2021, Tulala.2018, Huang.2018b, Li.2020, Mucllari.2023, Tamrin.2019, Mahyar.2022, Liu.2020, Donahue.2019, Li.2020b, Di.2019, Huang.2018}, and domain transfer \cite{Alawieh.2019, Yu.2021, Branikas.2023}.

Another observation is that the success factor lies in the similarity between domains and tasks reported within the literature. With more evidence showcasing the performance of \gls*{GenAI} in similar domains or tasks, the higher the likelihood of success in the target domain or task. Especially, the availability of a theoretical fundamental background seems to be a deciding factor for the use of \gls*{GenAI} in their work \cite{Alawieh.2021, Kim.2021, Meister.2021, Hartung.2022, Du.2023}.

\paragraph{Model Performance Characteristics.} 
However, a similar domain or task, does not guarantee a successful application in the target domain. Usually, model infrastructure and training logic adjustments are necessary. Therefore, \gls*{GenAI} architectures with lower complexity, and stable training were preferred over state-of-the-art architectures, with lots of hyperparameters \cite{Alawieh.2021, Yang.2022, Ye.2022, Kim.2021, Shen.2020, Meister.2021, Lv.2019, Du.2023}. Simpler models with smaller model sizes also increase inference speed, which plays a relevant role in industrial applications \cite{Alawieh.2019, Trent.2023}.

\paragraph{Data Requirements.}
The success of transferring a model architecture from a source domain to a target domain is mainly dependent on the similarity between the domains.
Domain similarity most commonly refers to the similarity of the underlying data distributions. Thus, data-related requirements regarding the data used were mentioned most frequently:
\begin{enumerate}
    \item \textbf{Data amount.} It is well known, that successfully training \gls*{GenAI} models requires a sufficient amount of data. Thus, it was of no surprise that the majority of publications reported that large amounts of data are needed. However, on the contrary, some authors also reported, that their proposed \gls*{GenAI} solution requires low amount of data (i.e. less than 100 samples) to effectively successfully generate realistic samples \cite{Eastwood.2021, Sundarrajan.2023, Seo.2020}.
    
    \item \textbf{Data diversity.} Modelling the underlying distribution of data requires that the samples for training cover a sufficient representation of the true data distribution. This usually becomes an issue if within the data structure, one class of data is highly over-represented, leading to more frequently generated samples from that class. Within our literature review, it could be observed that most commonly samples from defective manufactured products are missing, due to their naturally reduced availability during manufacturing \cite{Li.2023, Du.2023, Alam.2023}. To circumvent an uneven data diversity, it is possible to artificially generate more samples \cite{Posilovic.2022, Karamov.2021}.
          
    \item \textbf{Preprocessing.} High data quality is an important aspect in almost all machine learning approaches, especially the removal of noise during preprocessing. Although only occasionally mentioned, data cleaning \cite{Mahyar.2022} and preprocessing \cite{Alawieh.2019, Nguyen.2022, Donahue.2019} are extensive, oftentimes manual processes. If not applied properly, the noise of industrial machine vision applications generated during data acquisition can have an impact on model performance. By failing to remove the noise from the data, the model learns to replicate the noise together with the data.
    
    \item \textbf{Image Pairing.} In the special case of style transfer, a pair of images with a different style are required. Example work that require image pairing are \cite{Panda.2019, Nagorny.2018, Zheng.2023, Baranwal.2019}. Image pairing to a certain degree is comparable to labelling the data by separating the distribution manually into the different styles, thus reducing the distribution's complexity and therefore the necessary training effort. 
\end{enumerate}

\paragraph{Application Challenges.} 
Most authors reported application challenges of \gls*{GenAI} when the previously mentioned properties are not fulfilled. These resulting effects are poor-quality of generated images \cite{Holscher.2022}, like blurry images \cite{Liu.2021, Schaaf.2022}, or mode collapse \cite{Du.2023}, where samples of the generator cover a limited part of the source data distribution. Furthermore, \gls*{GenAI} methods had difficulties in sampling images out of distribution \cite{Singh.2022b}. Noise in training data also negatively impacts training performance. The poor image quality resulted in poor initial computer vision performance (e.g. data augmentation for defect detection), where relevant features were not learned by the model and therefore not generated \cite{Chen.2022, Liu.2021}. Applying traditional \gls*{GenAI} methods is usually limited in their ability to semantically control the image generation outcome \cite{Peres.2021}. Although existing solutions are capable in specifying the generation output, their variety is limited by either data availability or model architecture design.

Besides image quality, training instability was also reported \cite{Xie.2018}; Especially training convergence was difficult to achieve \cite{Schmedemann.2023, Alawieh.2021}. Simple variations like perspective shifts lead to poor sampling performance \cite{Maack.2022}. In occasional cases, the authors reported exploding gradients \cite{Tamrin.2019}. Apart from training instabilities, insufficient available hardware resulted in slow model training \cite{Noraas.2019, He.2023, Zhang.2022}. 



\subsection{Application of Generative Artificial Intelligence for Industrial Machine Vision Tasks}
\label{subsec:machine_vision_tasks}
The third research question aims to identify where \gls*{GenAI} has proven successful in industrial machine vision tasks. A successful application is assumed when the work was published in a peer-reviewed journal or conference proceeding. Firstly, the purpose of \gls*{GenAI} was analyzed and secondly, the machine vision tasks it was applied along with industrial domains. 

\begin{figure}[H]
    \centering
    \includegraphics[width=\textwidth]{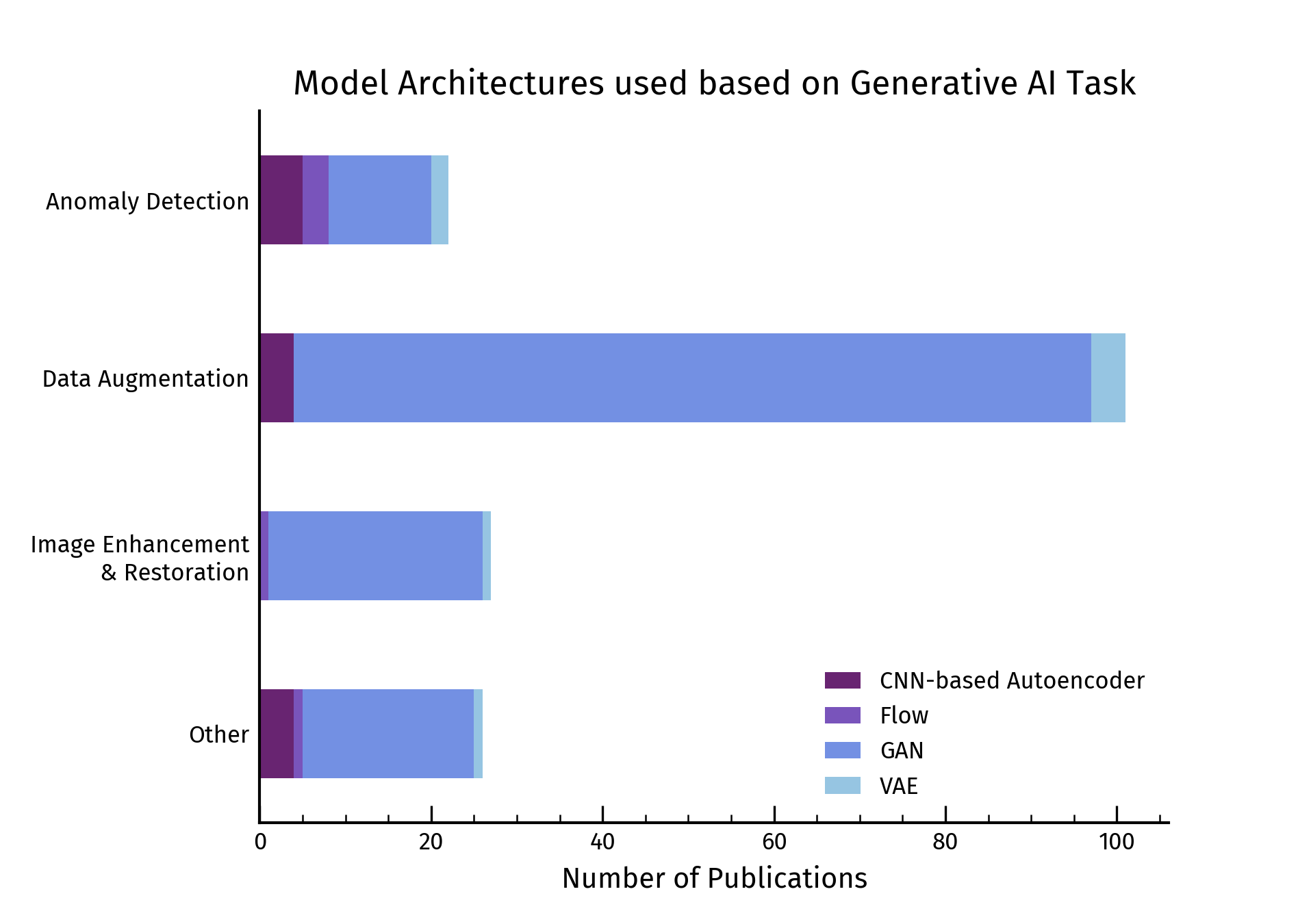}
    \caption[Model Architectures used based on \gls*{GenAI} Task]{Model Architectures used based on \gls*{GenAI} Task}\label{fig:chap:2:literature_analysis:genai_task}
\end{figure}

\paragraph{Generative Artificial Intelligence Tasks}
\figreft{fig:chap:2:literature_analysis:genai_task}
shows the correlation between the \gls*{GenAI} applications and the used \gls*{GenAI} model architectures. Four clusters were identified, for which most frequently \glspl{GAN} were utilized. The identified clusters include data augmentation, image enhancement, and segmentation, with some papers not fitting into any of these categories and falling into a fourth category labelled as others. The majority of publications address data augmentation, where the generated samples $\Tilde{x} \sim p_\mathcal{G} \li x \re$ are used to enrich the training data to cover more samples from the data distribution. Examples of image enhancement include use cases like image denoising, whereas image restoration include use cases like image inpainting. For this category, the idea lies in using the generative capabilities of \gls*{GenAI} models to propose a solution to repair or improve the corrupted image, conditioned on uncorrupted data. For anomaly detection, the general idea lies in first learning the data distribution, that estimates the likelihood of a sample belonging to the training data and afterward use that to detect anomalous samples. 
In the following, we address each application field with examples and challenges in the corresponding literature.

Data augmentation addresses the typical challenge regarding data scarcity and imbalance encountered in typical industrial machine vision problems. Manufacturing processes are often already optimized, thus, defects or some variants occur less frequent than defect-free variants. This results in imbalanced datasets used for training data-driven methods such as \gls*{DL} models. Also, in case large amounts of images are available, the effort for acquiring high-quality annotations lies oftentimes under economic constraints. 

The image generation capabilities of \gls*{GenAI} can be used to realize \textit{Data Augmentation}. \textsc{Na et al.} use a BigGAN architecture for conditional generation of scanning electron microscopy images of laser-processed surfaces~\cite{Na.2022}. They show that a small dataset of these images acquired with different process parameters is sufficient to generate surfaces with desired physical properties. A second non-generative network can be trained on synthetic data to predict physical properties such as the reflectance of real and generated surfaces. \textsc{Eastwood et al.} use a ProGAN for simulating high-resolution surface textures for Additive Manufacturing and coated surfaces~\cite{Eastwood.2022}. They extend the model to enable the conditional generation of surfaces with different texture categories and show that the approach can synthesize surfaces with known quantitative properties. In their outlook, they refer to the ability of semantic control given by representations of the latent space to further enhance their work. Both works indicate that \gls*{GenAI} can successfully learn the semantics of industrial machine vision domains and subsequently generate images considering the physical properties of the underlying application.   

Besides \textit{Data Augmentation}, \gls*{GenAI} is used for \textit{Image Enhancement/Restoration}, where the resolution or contrast of images can be enhanced or denoising can be applied~\cite{Lu.2021}. An example in the field of metrology is given by \textsc{Karamov et al}. They explore \gls*{GAN}-based inpainting for micro-CT images with the aim to apply it for reconstruction and CT artifact correction \cite{Karamov.2021}. Also in their work, the ability of \gls*{GAN} to learn the physical properties of the underlying use cases can be affirmed.  

The third cluster, \textit{Anomaly detection}, reformulates the need for detecting rare defects, which are either unknown or not well-defined~\cite{Lai.2018}, into a binary classification task, which predicts whether a given datum is normal or anomalous~\cite{Rippel.2020}. \textsc{Lai et al.} use an architecture based on a DCGAN for anomaly detection on industrial datasets~\cite{Lai.2018}. They use the latent space of a DCGAN trained on qualified samples only for anomaly detection because it contains all relevant information on the properties of qualified samples~\cite[p.~1446]{Lai.2018}. This suggests that the information encoded in the latent space can support decisions in industrial machine vision use cases.

Lastly, within the other categories, a variety of purposes for \gls*{GenAI} was investigated. For example, 
\cite{Yu.2021} presents a \gls*{GAN} architecture that guides domain adaption between features extracted with a \gls*{CNN} and the underlying data. Overall, \gls*{GenAI} has been applied successfully for different industrial machine vision purposes, addressing challenges such as poor resolution through image enhancement and data scarcity via augmentation techniques.

\paragraph{Machine Vision Tasks}



With the purposes of \gls*{GenAI} in industrial machine vision identified, their respective machine vision tasks are reviewed. 
The tasks identified include classification, object detection, semantic segmentation, and pose estimation. The papers indicate that classification and semantic segmentation are primarily used for detecting defects in manufacturing environments, while pose estimation mainly addresses the localization of manufacturing goods. Object detection is utilized for both defect detection in objects and general object identification. Machine vision tasks that lie outside the previously mentioned ones are classified as further. Publications that do not explain for what the captured images are initially used for are classified as not specified. In the following, for each machine vision task, \gls*{GenAI} applications are listed classified according to their purpose. 

Classification was the dominant machine vision task in the papers and dealt with categorizing images into predefined classes based on their visual content. Table \ref{table:papers_Classification} lists all papers for classification along with the purpose of the used \gls*{GenAI}. Further, many papers used \gls*{GenAI} for object detection tasks, which focuses on identifying and pinpointing specific objects within an image by drawing bounding boxes around them and classifying each object. Papers for object detection with the corresponding \gls*{GenAI} purpose are shown in Table \ref{table:papers_Object Detection}.

\renewcommand{\arraystretch}{1.4}
\begin{table}[htbp]
\centering
\caption{Papers found for machine vision task: classification}
\label{table:papers_Classification}
\begin{tabular}{p{0.25\textwidth}p{0.7\textwidth}}
\toprule
Purpose of \gls*{GenAI} & References \\
\midrule
Anomaly Detection & Kim et al. \cite{Kim.2021}, Xie et al. \cite{Xie.2021}, Schmedemann et al. \cite{Schmedemann.2023}, Hida et al. \cite{Hida.2021}, Balzategui et al. \cite{Balzategui.2021}, Mumbelli et al. \cite{Mumbelli.2023}, Lei et al. \cite{Lei.2021}, Tonnaer et al. \cite{Tonnaer.2019}, Wagner et al. \cite{Wagner.2019} \\
 Data Augmentation & Alawieh et al. \cite{Alawieh.2021}, Gao et al. \cite{Gao.2023}, Xu et al. \cite{Xu.2023}, Yang et al. \cite{Yang.2022}, Al Hasan et al. \cite{AlHasan.2021}, Byun et al. \cite{Byun.2022}, Ziabari et al. \cite{Ziabari.2022}, Wang et al. \cite{Wang.2023}, Chen et al. \cite{Chen.2022}, Meister et al. \cite{Meister.2021}, Yun et al. \cite{Yun.2020}, Oh et al. \cite{Oh.2020}, Liu et al. \cite{Liu.2021}, Li et al. \cite{Li.2023c}, Zhou et al. \cite{Zhou.2022}, Ross et al. \cite{Ross.2023}, Jin et al. \cite{Jin.2023}, He et al. \cite{He.2023}, Eastwood et al. \cite{Eastwood.2022}, Schaaf et al. \cite{Schaaf.2022}, Yi et al. \cite{Yi.2023}, Xie et al. \cite{Xie.2018}, Li et al. \cite{Li.2022b}, Heo et al. \cite{Heo.2020}, Guo et al. \cite{Guo.2020}, Lu et al. \cite{Lu.2020}, Shon et al. \cite{Shon.2021}, Zhang et al. \cite{Zhang.2022c}, Du et al. \cite{Du.2023}, Sundarrajan et al. \cite{Sundarrajan.2023}, Alam et al. \cite{Alam.2023}, Chung et al. \cite{Chung.2023}, Yang et al. \cite{Yang.2023}, Yang et al. \cite{Yang.2023b}, Zhang et al. \cite{Zhang.2019}, Niu et al. \cite{Niu.2020b}, Seo et al. \cite{Seo.2020}, Song et al. \cite{Song.2022}, Di et al. \cite{Di.2019}, Huang et al. \cite{Huang.2018} \\
 Image Enhancement \& Restoration & Wang et al. \cite{Wang.2021b}, Lu et al. \cite{Lu.2021}, Monday et al. \cite{Monday.2022}, Guo et al. \cite{Guo.2023}, Singh et al. \cite{Singh.2022}, Zhu et al. \cite{Zhu.2022b}, Li et al. \cite{Li.2020b}, Wei et al. \cite{Wei.2020}, Feng et al. \cite{Feng.2020}, Courtier et al. \cite{Courtier.2023}, Déau et al. \cite{Deau.2023}, Liu et al. \cite{Liu.2020b} \\
 Other & Pandiyan et al. \cite{Pandiyan.2022}, Noraas et al. \cite{Noraas.2019}, Yu et al. \cite{Yu.2021}, Lin et al. \cite{Lin.2021}, Wolfschläger et al. \cite{Wolfschlager.2023} \\
\bottomrule
\end{tabular}
\end{table}

\begin{table}[htbp]
\centering
\caption{Papers found for machine vision task: object detection}
\label{table:papers_Object Detection}
\begin{tabular}{p{0.25\textwidth}p{0.7\textwidth}}
\toprule
Purpose of \gls*{GenAI} & References \\
\midrule
 Anomaly Detection & Kuang et al. \cite{Kuang.2022}, Lai et al. \cite{Lai.2018}, Shen et al. \cite{Shen.2020}, Chen et al. \cite{Chen.2023}, Zhang et al. \cite{Zhang.2023}, Oz et al. \cite{Oz.2021} \\
 Data Augmentation & Li et al. \cite{Li.2023}, Ye et al. \cite{Ye.2022}, Zhang et al. \cite{Zhang.2022b}, Mao et al. \cite{Mao.2022}, Matuszczyk et al. \cite{Matuszczyk.2022}, Liu et al. \cite{Liu.2021b}, Mery et al. \cite{Mery.2020}, Zhu et al. \cite{Zhu.2022}, Li et al. \cite{Li.2024}, Rippel et al. \cite{Rippel.2020}, Li et al. \cite{Li.2022c}, Zhao et al. \cite{Zhao.2023}, Jin et al. \cite{Jin.2022}, Liu et al. \cite{Liu.2020}, Shirazi et al. \cite{Shirazi.2021}, Yin et al. \cite{Yin.2021}, Peres et al. \cite{Peres.2021}, Zheng et al. \cite{Zheng.2022}, Lv et al. \cite{Lv.2019}, Wen et al. \cite{Wen.2022}, Moriz et al. \cite{Moriz.2022}, Cannizzaro et al. \cite{Cannizzaro.2022}, Andrade et al. \cite{Barreiro.2020}, Wu et al. \cite{Wu.2021}, Niu et al. \cite{Niu.2020} \\
 Image Enhancement \& Restoration & Song et al. \cite{Song.2021}, Singh et al. \cite{Singh.2022b}, Wang et al. \cite{Wang.2021c}, Wang et al. \cite{Wang.2020}, Tang et al. \cite{Tang.2020} \\
 Other & Zheng et al. \cite{Zheng.2023} \\
\bottomrule
\end{tabular}
\end{table}

Moreover, a cluster dealing with semantic segmentation was identified. Semantic segmentation involves the division of images into regions or segments, unlike classification, which labels the entire image with a category, it labels each pixel to classify the entire image at a granular level, distinguishing different regions or objects with boundaries. Table \ref{table:papers_Segmentation} details all papers with semantic segmentation. The smallest cluster identified is pose estimation, which involves determining the orientation and position of an object or body within an image. In contrast, object detection identifies and locates objects but does not specifically assess their pose or orientation. All articles are listed in Table
\ref{table:papers_Pose Estimation}.

\begin{table}[htbp]
\centering
\caption{Papers found for machine vision task: segmentation}
\label{table:papers_Segmentation}
\begin{tabular}{p{0.25\textwidth}p{0.7\textwidth}}
\toprule
Purpose of \gls*{GenAI} & References \\
\midrule
Anomaly Detection & Zhang et al. \cite{Zhang.2024}, Shao et al. \cite{Shao.2022}, Maack et al. \cite{Maack.2022}, Lee et al. \cite{Lee.2023}, Park et al. \cite{Park.2022b}, Rudolph et al. \cite{Rudolph.2021} \\
Data Augmentation & Tang et al. \cite{Tang.2023}, Niu et al. \cite{Niu.2022}, Wei et al. \cite{Wei.2021}, Liu et al. \cite{Liu.2022b}, Li et al. \cite{Li.2020}, Kim et al. \cite{Kim.2020}, Lutz et al. \cite{Lutz.2021}, Liu et al. \cite{Liu.2019}, Yang et al. \cite{Yang.2019}, Donahue et al. \cite{Donahue.2019}, Liu et al. \cite{Liu.2023}, Yang et al. \cite{Yang.2024}, Niu et al. \cite{Niu.2022b}, Branikas et al. \cite{Branikas.2023}, Hedrich et al. \cite{Hedrich.2022}, Mertes et al. \cite{Mertes.2020} \\
Image Enhancement \& Restoration & Cheng et al. \cite{Cheng.2022}, Nguyen et al. \cite{Nguyen.2022}, Zhang et al. \cite{Zhang.2021} \\
Other & Panda et al. \cite{Panda.2019} \\
\bottomrule
\end{tabular}
\end{table}

\begin{table}[htbp]
\centering
\caption{Papers found for machine vision task: pose estimation}
\label{table:papers_Pose Estimation}
\begin{tabular}{p{0.25\textwidth}p{0.7\textwidth}}
\toprule
 Purpose of \gls*{GenAI} & References \\
\midrule
Data Augmentation & Park et al. \cite{Park.2022} \\
Image Enhancement \& Restoration & Yoon et al. \cite{Yoon.2023} \\
\bottomrule
\end{tabular}
\end{table}

The remaining articles were either clustered to further machine vision tasks, such as edge detection, or to unspecified tasks. These are listed in Table 
\ref{table:papers_Further} for Further tasks and Table
\ref{table:papers_Not Specified} for unspecified tasks.

\begin{table}[htbp]
\centering
\caption{Papers found for machine vision task: further}
\label{table:papers_Further}
\begin{tabular}{p{0.25\textwidth}p{0.7\textwidth}}
\toprule
Purpose of \gls*{GenAI} & References \\
\midrule
Data Augmentation & Na et al. \cite{Na.2022}, Huang et al. \cite{Huang.2018b}, Li et al. \cite{Li.2023b}, Kampker et al. \cite{Kampker.2023}, Cheng et al. \cite{Cheng.2022b}, Zhang et al. \cite{Zhang.2023b}, Wu et al. \cite{Wu.2023} \\
Image Enhancement \& Restoration & Panda et al. \cite{Panda.2022}, Wang et al. \cite{Wang.2021d}, Karamov et al. \cite{Karamov.2021}, Dong et al. \cite{Dong.2022} \\
Other & Nagorny et al. \cite{Nagorny.2018}, Hartung et al. \cite{Hartung.2022}, Tulala et al. \cite{Tulala.2018}, Alawieh et al. \cite{Alawieh.2019}, Trent et al. \cite{Trent.2023}, Mucllari et al. \cite{Mucllari.2023}, Mahyar et al. \cite{Mahyar.2022}, Liu et al. \cite{Liu.2022c}, Hoq et al. \cite{Hoq.2023}, Zhang et al. \cite{Zhang.2022}, Schmitt et al. \cite{Schmitt.2022}, Cao et al. \cite{Cao.2021} \\
\bottomrule
\end{tabular}
\end{table}

\begin{table}[htbp]
\centering
\caption{Papers found for machine vision Task: not specified}
\label{table:papers_Not Specified}
\begin{tabular}{p{0.25\textwidth}p{0.7\textwidth}}
\toprule
Purpose of \gls*{GenAI} & References \\
\midrule
Data Augmentation & Tan et al. \cite{Tan.2023}, Eastwood et al. \cite{Eastwood.2021}, Lin et al. \cite{Lin.2019}, Posilović et al. \cite{Posilovic.2022}, Tamrin et al. \cite{Tamrin.2019}, Hölscher et al. \cite{Holscher.2022}, Gobert et al. \cite{Gobert.2019}, Cha et al. \cite{Cha.2020}, Baranwal et al. \cite{Baranwal.2019} \\
Image Enhancement \& Restoration & Deepak et al. \cite{Deepak.2021}, Krishna et al. \cite{Krishna.2023} \\
Other & Guo et al. \cite{Guo.2023b}, Ramlatchan et al. \cite{Ramlatchan.2022}, Posilovic et al. \cite{Posilovic.2021} \\
\bottomrule
\end{tabular}
\end{table}






\section{Discussion and Conclusion}
\label{sec:discussion}
Research in \gls*{GenAI} has gathered significant attention for its potential in industrial domains. This review aimed to explore which architectures are used, which application challenges and requirements exist for enabling a successful application, and for which machine vision task \gls*{GenAI} is used for. From the review, the increase in research interest of \gls*{GenAI} in machine vision applications became apparent. With the predefined search string and study selection process, it is not guaranteed, that all relevant publications are covered in this review. Nonetheless, noticeable trends were successfully extracted. 

Research question 1 revealed that the majority of \gls*{GenAI} applications use \glspl*{GAN} as their architecture of choice. Due to this imbalance, a further division of \glspl*{GAN} into sub architectures, lead to countless \gls*{GAN} variations due to individual adjustments of the authors. It is fair to say, that the presented distinction of \gls*{GAN} architectures into their specific sub categories, is strongly debatable with multiple possible allocation solutions. The main issue lies in the fact that \gls*{GAN} architectures are not characterized by a single distinct feature, rather an accumulated number of feature gathered from previously proposed \glspl*{GAN}. Although a clear separation of \gls*{GAN} architectures could not be guaranteed, from the allocation a rough trend of \glspl*{GAN} could be observed. 

Research question 2 highlighted various challenges in the transferability of \gls*{GenAI} to industrial machine vision, such as data availability, preprocessing requirements and model architecture design choices. For this review, an industrial use case was assumed when the dataset was acquired in an industrial setting. Further investigation on how \gls*{GenAI} could be used outside academic environments, may reveal more insights into applying \gls*{GenAI} in industrial environments. 
Notably, only 15 articles (8.9 \%) had at least one industrial author. Therefore, it is important to acknowledge that most applications were found in research settings without direct industrial collaboration, which may indicate further requirements and properties from an industrial or economical perspective. Nonetheless, from a purely technical point of view, an overview about application challenges and data requirements was analyzed and presented, to support the evaluation of use cases regarding suitability for applying \gls*{GenAI}. 

Research question 3 demonstrated the diverse categories of \gls*{GenAI} for industrial machine vision tasks, that indicated major use of classification and object detection for all industrial domains. However, it is important to note that some authors do not explicitly specify the machine vision tasks for which the data was collected in the first place. Additionally, due to the usage of different terminology like "fault detection", which could refer to classification or object detection, a distinct classification of machine vision task was not always possible. 

Although \gls*{GenAI} emerged as a new research field for industrial machine vision, focusing on generating synthetic data, enhancing pattern recognition, and more, there was a lack of literature reviews addressing the various approaches and subfields within the research community. A \gls*{PRISMA} literature review was conducted to analyze \gls*{GenAI} for industrial machine vision to answer research questions about the \gls*{GenAI} architectures used, their requirements and properties in this domain, as well as successful applications in different machine vision tasks. The main findings indicate (i) the dominant use of GANs and \glspl*{VAE} as architectures, (ii) challenges related to the variety or shortage of image data, and (iii) diverse applications across different industrial machine vision tasks. However, with the ever-increasing number of publications in this research areas, the findings remain limited to the selected search string and depict only an incomplete snapshot of the research landscape. Nonetheless, this article provides a robust foundation for exploring literature in \gls*{GenAI} for industrial machine vision applications and gives future research directions as the field continues to evolve.

\section*{Declarations}

\subsection*{Funding}
Funded by the Deutsche Forschungsgemeinschaft (DFG, German Research Foundation) under Germanys Excellence Strategy - EXC-2023 Internet of Production – 390621612.  Funded by the Deutsche Forschungsgemeinschaft (DFG, German Research Foundation) - 507911127. Funded by the Ministry of Science and Education (BMBF) with the project WestAI – AI Service Center West under grant id 01IS22094D. This study was financed in part by the
Coordenação de Aperfeiçoamento de Pessoal de Nível Superior - Brasil (CAPES) - Finance Code 001.

\subsection*{Competing interests}
The authors have no competing interests to declare that are relevant to the content of this article.

\subsection*{Data Availability}
\label{sec:Data Availability}
Data will be made available on request.

\subsection*{Author Contributions}
Dominik Wolfschl\"ager and Hans Aoyang Zhou developed the idea for the review paper.
Dominik Wolfschl\"ager, Hans Aoyang Zhou, Constantinos Florides, Jonas Werheid, Hannes Behnen, Jan-Henrik Woltersmann and Tiago Pinto performed the literature search and data analysis.
The first draft of the manuscript was written by Dominik Wolfschl\"ager, Hans Aoyang Zhou, Constantinos Florides, Jonas Werheid, Hannes Behnen, Jan-Henrik Woltersmann and Tiago Pinto.
All authors critically revised the first draft.
All authors read and approved the final manuscript.

\bibliography{Manuscript}

\end{document}